\documentclass{article}
\usepackage{spconf,amsmath,graphicx,booktabs, multirow, rotating, breqn, dsfont, tabularx, adjustbox}

\title{Pseudo-Label Correction for Instance-Dependent Noise Using Teacher-Student Framework}
\name{Eugene Kim
\thanks{The author would like to thank Yifan Wu and Zhiting Hu from the University of California, San Diego for their helpful discussions and suggestions on this work.}}
\address{University of California, San Diego}
\begin{document}
\ninept
\maketitle
\begin{abstract}
The high capacity of deep learning models to learn complex patterns poses a significant challenge when confronted with label noise. The inability to differentiate clean and noisy labels ultimately results in poor generalization. We approach this problem by reassigning the label for each image using a new teacher-student based framework termed P-LC (pseudo-label correction). Traditional teacher-student networks are composed of teacher and student classifiers for knowledge distillation. In our novel approach, we reconfigure the teacher network into a triple encoder, leveraging the triplet loss to establish a pseudo-label correction system. As the student generates pseudo labels for a set of given images, the teacher learns to choose between the initially assigned labels and the pseudo labels. Experiments on MNIST, Fashion-MNIST, and SVHN demonstrate P-LC's superior performance over existing state-of-the-art methods across all noise levels, most notably in high noise. In addition, we introduce a noise level estimation to help assess model performance and inform the need for additional data cleaning procedures.

\end{abstract}
\section{Introduction}
\label{sec:intro}

As the size of training dataset increases, generalization of deep neural networks (DNNs) is expected to improve due to their efficient pattern memorization capabilities \cite{zhang2021understanding}. However, the ability to memorize complex patterns can lead to worse generalization performance depending on the accuracy of the dataset annotation. DNNs struggle to distinguish noise from clean data, which results in overfitting on noise and poor generalization \cite{frenay2013classification, sukhbaatar2015training}. While accurately labeled data is essential, the majority of data label collection methods, e.g., crowd-sourcing and web crawling, are expensive and susceptible to mistakes, particularly when dealing with large-scale datasets \cite{welinder2010online,raghavan2001crawling,fergus2010learning}. Hence, the research area of weakly-supervised learning holds much importance as it aims to improve model robustness in the presence of partially, imprecisely, or inaccurately labeled data \cite{zhou2018brief}. In our paper, we tackle the challenge of learning with noisy labels.

To establish proper benchmarks, researchers have introduced various forms of controlled synthetic noise to replicate real-world noise conditions \cite{angluin1988learning, xia2020part, chen2021beyond, scott2013classification}. The two most basic forms include symmetric noise and class-conditional noise (CCN). In a symmetric noise distribution, the corruption rate for all images are independently and identically distributed \cite{angluin1988learning}, whereas the corruption probability under a CCN assumption depends on the specific class of the image \cite{scott2013classification}. Nevertheless, in practice, the corruption probability for each image tends to differ regardless of its class association \cite{jiang2020beyond}. In order to more closely follow the real-world noise distribution, we adopt an approach involving convolutional neural networks (CNN) to generate instance-dependent noise (IDN) \cite{chen2021beyond}.

Earlier research in developing robust methods for noisy labels have leveraged statistical learning techniques, while more recent techniques have incorporated deep learning approaches. The statistical learning methods primarily encompass two forms: surrogate loss and noise rate estimation. For the surrogate loss approach, Masnadi-Shirazi et al. introduced SavageBoost, a boosting algorithm derived from a robust non-convex loss for corrupted binary classification \cite{masnadi2008design}. Patrini et al. proposed a loss correction method using at most a matrix inversion and multiplication, but relied on the assumption that each class corruption probability is known \cite{patrini2017making}. For noise rate estimation methods, Menon et al. introduced a class-probability estimator by optimizing balanced error and AUC \cite{menon2015learning}. Both types of past statistical learning methods heavily relied on impractical noise assumptions, e.g. known class noise distribution, or were limited to binary classification, making them less effective for real-world scenarios.

More recent works have incorporated deep learning methods to correct or reweight weakly labeled data to achieve state-of-the-art (SOTA) classification accuracy \cite{zheng2019meta,liu2015classification, shu2019meta}. Ren et al. used meta-learning to adjust the weights of training examples based on the gradient directions \cite{ren2018learning}. Li et al. trained the teacher classifier on a small and clean sample, while leveraging a Wikipedia-based knowledge graph to guide the training process of a student classifier \cite{li2017learning}. To achieve competitive performance, this method required a carefully designed knowledge graph harvested from an external source of data. Our proposed method eliminates the need for additional data gathering and processing.

In contrast to the conventional teacher-student framework, our teacher network operates as a correction system for the predictions generated by the student network. We deem our method as pseudo-label correction (P-LC). P-LC can be divided into two phases: the teacher-student training phase and the noise correction phase. During the teacher-student training phase, both networks are separately trained on a clean dataset with their corresponding losses. During the correction phase, the student's task is to create reliable pseudo labels, while the teacher decides whether the initially assigned labels or the pseudo labels are more fit for the given images. The student is then retrained on the corrected dataset and makes predictions for the test set.

Our simple yet highly effective technique comes with two main advantages: reduced overall training time and self-adjustment to noise levels. The teacher and student network are trained concurrently on the clean dataset. Note the clean dataset is much smaller than the noisy dataset, also helping with the training time. Moreover, P-LC adjusts to noise levels without explicit tuning, allowing for consistent performance across all noise levels. As noise increases, the teacher network progressively places more trust in the pseudo labels over the initially assigned labels. Based on this model behavior, we further the use case by estimating the noise level of the dataset. We calculate the proportion of different labels between the corrected and noisy dataset as the estimation for the noise level.

The key findings and contributions of this paper can be outlined as follows:
\begin{itemize}
  \item We introduce a novel teacher-student framework that leverages pseudo labels to correct potentially misleading noisy labels.
  \item To the best of our knowledge, we propose the first teacher-student based approach for noise level estimation.
  \item We conduct experiments on three common benchmark datasets with varying noise levels and the proposed method outperforms SOTA  methods, most visibly in high noise levels.
\end{itemize}





\section{Problem Formulation and Assumptions}
\label{sec:instance-dependent label noise assumption}

\subsection{Preliminary}
For a $K$-class classification problem, we define the feature space $X$ and label space $Y = \{1, \dots, k\}$. Consistent with prior research, we are given a noisy training dataset $D_{noise} = \{(x_{i}, y'_{i})\}_{i=1}^{n}$ and clean dataset $D_{clean} = \{(x_{i}, y_{i})\}_{i=1}^{m}$, where $m << n$ ($m$ is the size of $D_{clean}$ and $n$ is the size of $D_{noise}$) \cite{charikar2017learning, veit2017learning, zheng2019meta, li2017learning}. The corrected dataset after implementing our method is defined as $D_{corrected} = \{(x_{i}, \bar y_{i})\}_{i=1}^{n}$. Note $D_{clean}$ has a uniform label distribution, and therefore, does not guarantee the same label distribution as $D_{noise}$ or $D_{corrected}$.

\subsection{Noise Assumptions}

\begin{figure}
\centering
\includegraphics[width=6cm]{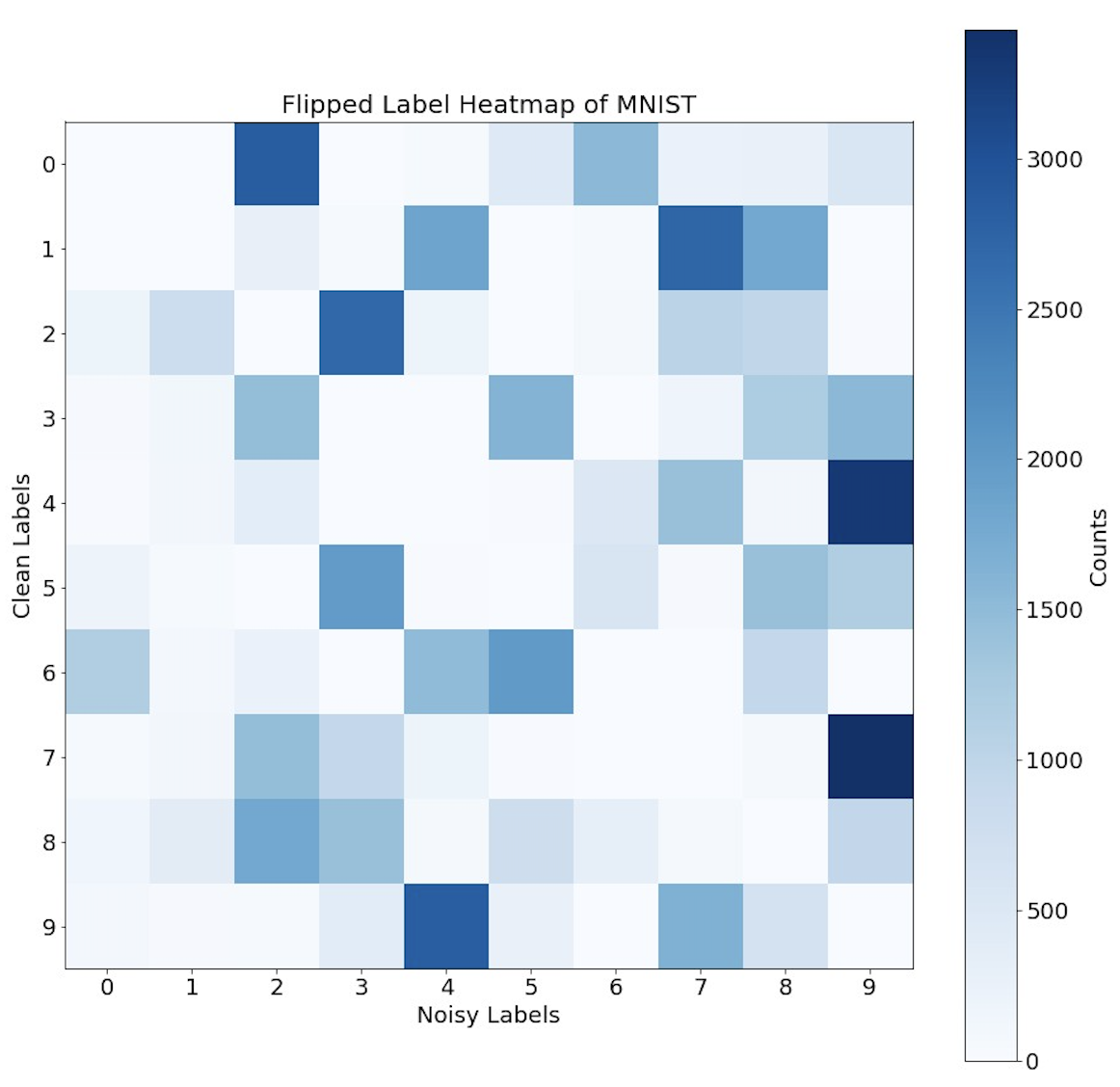}
\caption{Flipped label heatmap of MNIST under the IDN assumption.}
\label{fig_1}
\end{figure}
IDN with varying noise levels are injected into clean datasets to closely imitate real-world noise generation while retraining a controlled environment. We adopt an IDN generation approach that leverages the soft label predictions from a CNN trained on the entire clean dataset \cite{chen2021beyond}. For each input, the CNN's prediction with the highest likelihood is used as the noisy label. By enforcing a likelihood for each image, it goes beyond the class-conditional assumptions and satisfies the requirements for IDN.

The distribution of flipped labels based on the IDN assumption indicates that for most of classes, there exists a single corresponding class with the highest flip likelihood. Thus, the  majority of potential corruption classes do not provide any additional information. As seen in Fig. \ref{fig_1}, images depicting the number 4 are predominantly flipped to resemble the number 9, while images of the number 1 tend to be transformed into the number 4, and so forth. This observation forms the basis for choosing a hard over soft label correction method. Moreover, the performance of existing soft label correction methods drops significantly as the noise level increases \cite{chen2021beyond, zheng2019meta}. Recent study by Wei et al. reports that label smoothing (LS) actually decreases model performance in high noise settings \cite{wei2021smooth}. The computational cost of the soft label is also more expensive compare to the one-hot encoded label. Thus, we develop a simple yet highly effective approach by employing a hard pseudo-label correction technique.


\section{Methodology}
\label{sec:meta pseudo-label correction}

\subsection{Overview}
\label{ssec: overview}

Our proposed method P-LC involves a teacher-student network and can be divided into two separate phases: the teacher-student training phase and noise correction phase. During the teacher-student training phase, the teacher network learns to differentiate images from the same class (positive pairs) and images from different classes (negative pairs). The student network focuses on standard image classification. Both networks are trained on the small set of clean data $D_{clean}$. During the noise correction phase, the student network generates pseudo labels for all images in the noisy dataset $D_{noise}$. The teacher network then decides whether to use the pseudo labels or initially assigned labels as the new labels in the corrected dataset $D_{corrected}$. As a final step, the student network is retrained on $D_{corrected}$ to provide predictions for the test set.

\begin{figure*}[t]
\includegraphics[width=1\textwidth]{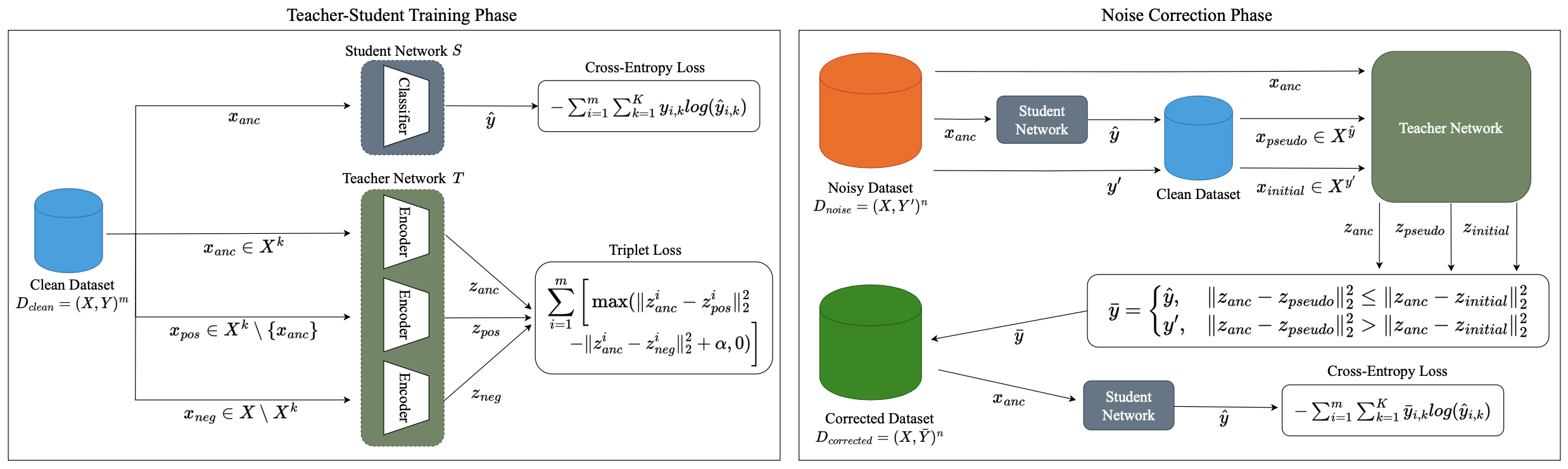}
\caption{P-LC's computation graph. The left panel shows the training process of the teacher-student network on the clean dataset. On the right panel, we show the noise correction process composed of pseudo-label generation, data sampling, and label correction.}
\label{fig_2}
\end{figure*}

\subsection{Teacher-Student Architecture}
\label{ssec:teacher-student training}

For the teacher network $T$, we use a Siamese framework composed of three identical encoders, all consisting of the same architecture and weights \cite{schroff2015facenet}. When training on the clean data $D_{clean}$, the inputs consist of an anchor image $x_{anc}$, a positive image $x_{pos}$, and a negative image $x_{neg}$ defined as follows
\begin{equation}
\begin{aligned}
    x_{anc} \in X^{k},  x_{pos} \in X^{k} \setminus \{x_{anc}\}, x_{neg} \in X \setminus X^{k}
\end{aligned}
\end{equation}
where $X^{k}$ refers to all images in feature space $X$ with class label $k$. This ensures the anchor image is from the same class as the positive image but different class as the negative image. Once $x_{anc}$ has a corresponding $x_{pos}$ and $x_{neg}$, the network encodes $x_{anc}$, $x_{pos}$, and $x_{neg}$ as $z_{anc}$ (anchor embedding), $z_{pos}$ (positive embedding), and $z_{neg}$ (negative embedding), respectively. For every anchor embedding, we want to decrease the distance from the positive embedding but increase the distance from the negative embedding as shown below
\begin{equation}
\begin{aligned}
\|z_{anc} - z_{pos}\|^{2}_{2} + \alpha < \|z_{anc} - z_{neg}\|^{2}_{2}
\end{aligned}
\end{equation}
where $\alpha$ is a hyperparameter that sets the minimum difference between the positive and negative embeddings. The teacher loss $\ell_{T}$ is then minimized, which is the triplet loss displayed below
\begin{equation}
\begin{aligned}
&\ell_{T} = \sum_{i=1}^{m}\max(\|z^{i}_{anc} - z^{i}_{pos}\|^{2}_{2} - \|z^{i}_{anc} - z^{i}_{neg}\|^{2}_{2} + \alpha, 0) \\
\end{aligned}
\end{equation}
For the student network $S$, we use a combination of CNN layers to learn lower level image embeddings and fully-connected layers to make label predictions. We minimize the student loss $\ell_{S}$, which is the cross-entropy loss displayed below
\begin{equation}
\begin{aligned}
    \ell_{S} = -\sum_{i=1}^{m} \sum_{k=1}^{K}y_{i,k} log(\hat{y}_{i,k})
\end{aligned}
\end{equation}
where $\hat{y}$ is the model prediction and $y$ is the true label. The teacher and student network can be trained concurrently on $D_{clean}$ because $\ell_{S}$ and $\ell_{T}$ rely on different inputs.

\subsection{Learning to Correct Pseudo-labels}
\label{ssec:learning to correct pseudo-labels}

Once both the teacher and student network are trained on $D_{clean}$, the noise correction phase begins. For the entire correction phase, every image in $D_{noise}$ is treated as an anchor image $x_{anc}$. The correction phase is composed of three main components: pseudo label generation, data sampling, and label correction.

For each input $x_{anc}$ in $D_{noise}$, the student network generates a pseudo label $\hat{y}$. $\hat{y}$, alongside the initial label $y'$, present two potential classes of $x_{anc}$. In the case that $\hat{y} = y'$, no correction is required by the teacher network. $x_{anc}$ retains $y'$ as the label in the corrected dataset $D_{corrected}$. Otherwise, we randomly sample two images $x_{pseudo}$ and $x_{initial}$ from $D_{clean}$ with different class labels (i.e. $x_{pseudo} \in X^{\hat y}$ and $x_{initial} \in X^{y'}$). The teacher network then encodes $x_{anc}$, $x_{pseudo}$, and $x_{initial}$ as $z_{anc}$, $z_{pseudo}$, and $z_{initial}$. Based on the dissimilarity correction metric, the corrected label $\bar{y}$ is computed as
\begin{equation}
\label{correction}
\begin{aligned}
& \bar{y} = \begin{cases} \hat{y}, & \|z_{anc} - z_{pseudo}\|^{2}_{2} \leq \|z_{anc} - z_{initial}\|^{2}_{2} \\ y',& \|z_{anc} - z_{pseudo}\|^{2}_{2} > \|z_{anc} - z_{initial}\|^{2}_{2}.
 \end{cases} \\
\end{aligned}
\end{equation}
\eqref{correction} indicates the teacher network assigns the label of the image that most closely resembles the anchor image. $(x_{anc}, \bar{y})$ is added to $D_{corrected}$. The student network retrains on $D_{corrected}$ and is evaluated on the test data. The complete process is shown in Fig. \ref{fig_2}.

We can improve the label correction accuracy by increasing the number of samples from $D_{clean}$. With a larger pool of images for comparison alongside $x_{anc}$, we reduce the variance and improve generalization as seen in previous studies \cite{sung2018learning, ravi2016optimization}. In practice, we sample five images instead of one image from $D_{clean}$. We then assign $x_{anc}$ to the label of the images with the highest number of $\bar{y}$ transformations. This approach is particularly effective when dealing with low-quality or corrupted sampled images.

P-LC has two main advantages over existing methods: reduced overall training time and self-adjustment to noise levels. First, the separate loss functions used in the teacher-student network enables concurrent training. Second, P-LC does not require label smoothing, reducing the computation cost when calculating $\ell_{S}$. Beyond reduced training time, P-LC can adjust to varying noise levels with a properly trained teacher-student network. In high noise levels, more label corrections are made by the teacher network due to the increased cases where $\hat{y} \neq y'$. This dynamic adjustment of label corrections based on noise level represents an internal tuning mechanism inherent to P-LC.

\begin{table*}
\caption{Classification accuracy on MNIST, Fashion-MNIST, and SVHN with different instance-dependent label noise levels.}
\renewcommand{\arraystretch}{1.5}
\centering
\resizebox{1.8\columnwidth}{!}{
\begin{tabular}{c|c|c|c|c||c|c|c|c||c|c|c|c} 
\hline
{Method} & \multicolumn{4}{c||}{MNIST} & \multicolumn{4}{c||}{Fashion-MNIST} & \multicolumn{4}{c}{SVHN}        \\ 
\cline{2-13}
                        &  IDN-20\%      &  IDN-30\%     &  IDN-40\%                      &  IDN-50\%      &  IDN-20\%      &  IDN-30\%                 &  IDN-40\% &  IDN-50\% &  IDN-20\%                  &   IDN-30\% &  IDN-40\% &  IDN-50\%           \\ 
\hline
SOTA                & 93.07 & 88.53 & 77.48                 & 73.27 & 82.84 & 82.15            & 79.98 & 72.96 & 	83.09 & 78.73 & 	73.64 & 64.90 \\ 
    & $\pm$ 0.22   &  $\pm$ 0.25  &  $\pm$ 0.79   &  $\pm$ 0.05  & $\pm$ 0.12 & $\pm$ 0.01  &  $\pm$ 0.19 & $\pm$ 0.37 & $\pm$ 0.27 & $\pm$ 0.74 & $\pm$ 0.54 & $\pm$ 0.53 \\
\hline
P-LC                  & \bf{94.81} & \bf{94.26} & \bf{93.71}                & \bf{93.07} & \bf{83.60} & \bf{82.61}            & \bf{80.89} & \bf{79.78} & \bf{83.31} & \bf{82.03} & \bf{78.62} & 	\bf{76.15}\\ 
& $\pm$ 0.20   &  $\pm$ 0.15   &  $\pm$ 0.25   &  $\pm$ 0.21  & $\pm$ 0.06  & $\pm$ 0.12  &  $\pm$ 0.14 & $\pm$ 0.42 & $\pm$ 0.13 & $\pm$ 0.13 & $\pm$ 0.37 & $\pm$ 0.11 \\
\hline 

\end{tabular}}
\label{results}
\end{table*}

\subsection{Noise Level Estimation}

We further our work by introducing a noise level estimator built upon the same teacher-student architecture. Once we obtain $D_{corrected}$ using P-LC, we can estimate the noise level with one additional calculation. We simply compute the proportion $p_{noise}$ of different labels between $D_{corrected}$ and $D_{noisy}$ as displayed below:
\begin{equation}
    \begin{aligned}
    p_{noise} = \frac{\sum_{i=1}^{n}I(\bar{y}_{i}, y'_{i})}{n},
    \end{aligned}
\end{equation}
where $I(\bar{y}_i, y_i') = \begin{cases} 1, & \bar{y}_{i} \neq y'_{i} \\ 0,& \mathrm{otherwise}. \end{cases}$Computing $p_{noise}$ provides benefits when the true noise level is both unknown and known. In the case of an unknown true noise level, an estimation can help assess model performance and inform the need for additional data cleaning procedures. In our study, we leverage $p_{noise}$ as an initial indicator to evaluate P-LC's relabeling accuracy. Depending on the proximity of $p_{noise}$ to the true noise level, we make adjustments to the hyperparameters of the teacher-student network. Table \ref{noise_estimation} presents the averaged noise level estimations across three datasets. Although P-LC struggles to estimate exact noise levels, it correctly ranks the noise level ranging from 20\% IDN to 50\% IDN.
\begin{table}
    \centering
    \caption{P-LC noise level estimations on MNIST, Fashion-MNIST, and SVHN}
    \vspace{10pt}
    \begin{tabular}{l@{\hspace{6pt}}ccc@{\hspace{6pt}}c}
        \toprule
         & IDN-20\% & IDN-30\% & IDN-40\% & IDN-50\% \\
        \midrule
        MNIST & 13.40  & 22.59 & 31.92 & 41.69 \\
             & $\pm$ 0.16 & $\pm$ 0.19 & $\pm$ 0.24 & $\pm$ 0.28 \\
        \midrule
        F-MNIST & 8.09 & 12.15 & 17.96 & 25.06 \\
             & $\pm$ 0.18 & $\pm$ 0.25 & $\pm$ 0.18 & $\pm$ 0.41 \\
        \midrule
        SVHN & 15.37 & 22.35 & 23.56 & 30.21 \\
             & $\pm$ 0.06 & $\pm$ 0.10 & $\pm$ 0.24 & $\pm$ 0.21 \\
        \bottomrule
    \end{tabular}
\label{noise_estimation}
\end{table}

\section{Experiments}
\label{sec:experiments}

\subsection{Datasets}
\label{ssec:datasets}
We evaluate P-LC against SOTA methods \cite{han2018co, thulasidasan2019combating, chen2021beyond} using three image recognition datasets: MNIST, Fashion-MNIST, and SVHN. For existing methods that do not require a clean dataset, we inject the training data with IDN at rates of 20\%, 30\%, 40\%, and 50\%. In contrast, for our method, we first construct the clean dataset $D_{clean}$ by sampling from the training set. The remaining training set is injected with four levels of IDN at rates of 22\%, 32\%, 42\%, and 52\% to compensate existing methods that do not require a clean dataset. We intentionally place our method at a disadvantage by slightly increasing the noise across all experiments. A summary of each dataset is provided in Table \ref{overview}.
\begin{table}[ht]
    \centering
    \caption{Overview of our datasets and teacher-student network architectures. Note the preceding numbers in CNN-4 and Siamese-5 refer to the total number of layers.}
    \vspace{10pt}
    \begin{tabular}{l@{\hspace{6pt}}|ccc}
        \toprule
        Dataset & MNIST & Fashion-MNIST & SVHN \\
        \midrule
        \# classes & 10 & 10 & 10 \\
        RGB & No & No & Yes \\
        Train & 60,000 & 60,000 & 73,257 \\
        Test & 10,000 & 10,000 & 26,032 \\
        \midrule
        Clean & 1,200 & 1,200 & 1,465 \\
        Noisy & 58,800 & 58,800 & 71,792 \\
        \midrule
        Teacher net. & Siamese-5 & Siamese-5 & Siamese-5 \\
        Student net. & CNN-4 & CNN-4 & ResNet-18 \\
        \bottomrule
    \end{tabular}
\label{overview}
\end{table}

\subsection{Implementation Details}
\label{ssec:implementation details}

We compare our method against four existing methods: cross-entropy (CE) loss, co-teaching \cite{han2018co}, deep abstaining classifier (DAC) \cite{thulasidasan2019combating}, and self-evolution average label (SEAL) \cite{chen2021beyond}. To ensure a fair and consistent comparison, all implementations are done on PyTorch and experiments are executed on NVIDIA GeForce GTX 1080 Ti. We use the same classifier architecture across all methods, i.e., ResNet-18 for SVHN and 4-layered CNN for MNIST and Fashion-MNIST. Existing methods are trained three times (seed 0, 1, and 2) for 50 epochs on every noise level. For our method, we use 50 epochs for retraining on the corrected dataset $D_{corrected}$. The batch size is set to 64. Due to limited space, we report the highest accuracy achieved among the four existing methods as SOTA in Table \ref{results}.

\subsection{Results}
\label{ssec:results}

In Table \ref{results}, we present the averaged accuracies across 4 distinct IDN levels, ranging from 20\% to 50\%. Based on our experimental results, P-LC consistently outperforms existing methods across all three datasets. The most notable differences are seen in high noise settings. While SOTA methods experience a significant decline in accuracy as the noise level increases, P-LC's test accuracy depreciates at a lower rate.

Several factors contribute to the reduced performance observed in SOTA methods at high noise levels. In the case of co-teaching, it relies on sampling clean images from noisy data to guide student training \cite{han2018co}. However, as noise increases, the pool of available clean images for selection diminishes. DAC is a different approach that abstains from making predictions when the class label's uncertainty is high \cite{thulasidasan2019combating}. Nonetheless, DAC faces the same limitations as co-teaching. High noise levels restrict the available information from the training data, resulting in poor generalization. We address this problem by employing a relabeling over reweighting technique. This enables the model to undergo training on a complete dataset, preventing the limitation of data as noise increases. SEAL, the third SOTA method, uses a label smoothing technique to iteratively re-adjust the label distribution of images \cite{chen2021beyond}. According to research conducted by Wei et al.,  label smoothing leads to reduced accuracy in high noise settings. \cite{wei2021smooth}. We overcome this issue by generating hard instead of soft-pseudo labels. By incorporating both hard-pseudo labels and relabeling techniques into P-LC, the proposed method outperforms SOTA methods across all noise levels.

\section{Conclusion}
\label{sec:conclusion}

In this paper, we address the issue of noisy labels from a label correction perspective. We explore the limitations of existing methods, specifically label smoothing and reweighting techniques, in the presence of instance-dependent noise. We then introduce a novel teacher-student framework designed to address these challenges by integrating hard-pseudo labels with label corrections. Essentially, the teacher encoder operates as a correction system for predictions made by the student classifier. P-LC comes with two main advantages: reduced overall training time and self-adjustment to noise levels. Empirical experiments on MNIST, Fashion-MNIST, and SVHN with varying noise levels demonstrate the superior noise robustness of P-LC compared to SOTA methods, particularly in high noise environments. Furthermore, we introduce a simple noise level estimation to help assess model performance and inform the need for additional data cleaning procedures.







\small
\bibliographystyle{IEEEbib}
\bibliography{strings,refs}

\end{document}